\documentclass[conference]{IEEEtran}
\IEEEoverridecommandlockouts
\usepackage{cite}
\usepackage{amsmath,amssymb,amsfonts}
\usepackage{algorithmic}
\usepackage{graphicx}
\usepackage{textcomp}
\usepackage{xcolor}
\usepackage{booktabs}
\usepackage{multirow}
\usepackage{float}

\usepackage[colorlinks=true, urlcolor=blue]{hyperref}
\def\BibTeX{{\rm B\kern-.05em{\sc i\kern-.025em b}\kern-.08em
    T\kern-.1667em\lower.7ex\hbox{E}\kern-.125emX}}
\begin{document}

\title{M$^2$OE$^2$-GL: A Family of Probabilistic Load Forecasters That Scales to Massive Customers 
\vspace{-5mm}
}

\author{\IEEEauthorblockN{Haoran Li, Zhe Cheng, Muhao Guo, Yang Weng}
\IEEEauthorblockA{\textit{Department of Electrical, Computer and Energy Engineering} \\
\textit{Arizona State University},
Tempe, USA \\
\mbox{\{lhaoran, zcheng55, mguo26, yang.weng\}@asu.edu}}
\and
\IEEEauthorblockN{Yannan Sun, Victor Tran, John Chainaranont}
\IEEEauthorblockA{\textit{Oncor Electric Delivery} \\
Dallas, TX \\
\mbox{\{Yannan.Sun, Victor.Tran, John.Chainaranont\}@oncor.com}}
}
\maketitle

\begin{abstract}
Probabilistic load forecasting is widely investigated, laying the foundations for power system planning, operation, and risk-aware decision-making. In particular, deep learning–based forecasters have shown a strong ability to capture complex temporal and contextual patterns, achieving substantial gains in accuracy. However, at the scale of thousands or even hundreds of thousands of loads in large distribution feeders, a fundamental deployment dilemma emerges: training and maintaining one model per customer load is computationally expensive and storage-intensive, while using a single global model for all loads ignores distributional shifts across customer types, locations, and phases. Existing work typically focuses on single-load forecasters, global models trained across multiple loads, or adaptive/personalized models designed for relatively small datasets, but rarely addresses the combined challenges of heterogeneity and scalability in large-scale feeders.
To tackle this problem, we propose M$^2$OE$^2$-GL, a global-to-local (GL) extension of a cutting-edge probabilistic forecaster (M$^2$OE$^2$). We first pre-train a single global M$^2$OE$^2$ base model across all feeder loads and then develop lightweight fine-tuning to derive a compact family of forecasters. We evaluate M$^2$OE$^2$-GL on realistic data from our industry partners, showing that the proposed method yields substantial error reductions, while remaining scalable to extensive loads. The source code is publicly available at \href{https://github.com/haorandd/M2oE2_load_forecast.git} {\textcolor{blue}{https://github.com/haorandd/M2oE2\_load\_forecast.git}}.
\end{abstract}


\section{Introduction}

Accurate load forecasting plays a vital role in reliable and economic power system estimation, planning, and operation \cite{han2021task,8973580,8784550,ref:Haoran2025L,10163982}. With the widespread deployment of smart meters, recent practice has shifted toward bottom-up, high-resolution forecasting for customer and transformer loads in distribution feeders \cite{li2025exarnn,li2025external}. In particular, probabilistic forecasting \cite{li2025external,wang2022novel,9321099} quantifies uncertainty and provides more informative estimates for disaggregated loads. However, these tasks remain challenging due to the intrinsic complexity of distribution-level demand \cite{wang2018combining,10138375}, which is strongly influenced by calendar effects, weather, electricity prices, and other exogenous factors.

Over the past decades, load forecasting models have evolved from classical statistical formulations to modern data-driven approaches. Early work relied on linear univariate models such as autoregressive integrated moving average (ARIMA) \cite{yildiz2017review}, exponential smoothing \cite{christiaanse2007short}, and state–space formulations \cite{de2022state}, which were later augmented by machine learning methods (e.g., support vector regression \cite{chen2004load}, gradient boosting \cite{taieb2014gradient}, random forests \cite{lahouar2015day}) to better capture nonlinearities. More recently, deep learning–based forecasters, such as recurrent neural networks (RNNs) \cite{shi2017deep}, long short-term memory (LSTM) networks \cite{kong2017short}, gated recurrent units (GRUs) \cite{abumohsen2023electrical}, convolutional neural networks (CNNs) \cite{rafi2021short}, and transformer-based architectures \cite{wang2022transformer}, have achieved state-of-the-art performance by learning complex temporal patterns from large-scale data. In parallel, the community has moved from purely univariate models based only on historical load to multivariate formulations \cite{haq2019new} that incorporate rich external covariates such as calendar indicators, weather variables, prices, and customer attributes, enabling more expressive spatio–temporal and context-aware forecasting. Building on this line, our recent Meta–Mixture-of-Experts for External Data (M$^2$OE$^2$) \cite{li2025external} is a state-of-the-art forecaster that uses a mixture-of-experts over external variables to smoothly modulate the parameter subspace of a base neural network, achieving strong predictive performance across diverse operating conditions while maintaining computational efficiency.

However, these models still lack the capacity to serve massive numbers of customer loads in large distribution systems. On the one hand, strong heterogeneity and distributional differences (see Fig. \ref{fig:Oncor_Transformer_vis}) across customers, phases, and locations mean that a single global model cannot fully capture the diverse patterns of individual or transformer-level loads. On the other hand, training and maintaining separate deep models for every customer quickly becomes infeasible for utilities due to scalability constraints in storage, computation, and model management.

To address these challenges, one natural direction is to consider adaptation methods that reuse common knowledge across groups of customers through lightweight adaptive operations, rather than training fully independent models for each load. Transfer learning adapts a model trained on a source domain to a single target domain \cite{ref:Haoran2022T,li2022domain,9338252,9805850}, but this one-to-one paradigm does not scale well when utilities must support many distinct customer groups and operating regimes. Meta-learning aims to learn good initializations or update rules across tasks \cite{talkhi2024using,talagala2023meta,abdallah2025evaluation}, enabling fast adaptation to new tasks, but typically relies on a two-stage inner–outer training procedure that is computationally expensive at large scale. Test-time adaptation updates models on-the-fly using only test-time data \cite{guo2025online}, which can be lightweight but generally offers weaker accuracy gains because rich offline data is not used. In contrast, pre-training and fine-tuning \cite{liang2025energygpt} provide a more practical and scalable solution in our setting: our Meta–Mixture-of-Experts for External Data (M$^2$OE$^2$) serves as a high-capacity pre-trained model that captures complex relationships between loads and external variables, and lightweight fine-tuning can then adapt this single global backbone to a unique customer groups, yielding an efficient and affordable adaptation strategy and storage requirements for utilities.

To this end, we propose M$^2$OE$^2$-GL, a global-to-local extension of M$^2$OE$^2$ designed to efficiently handle massive numbers of customer loads. The framework consists of three phases: (i) offline pre-training, where a single global M$^2$OE$^2$ model is learned from all customer groups, (ii) group-wise fine-tuning, where we adopt a low-rank adaptation (LoRA)–style strategy \cite{hu2022lora} and explicitly identify which subsets of parameters should be fine-tuned, providing both conceptual justification and numerical validation, and (iii) online prediction, where utilities deploy the frozen pre-trained backbone together with a compact set of stored additive parameters per group, enabling scalable real-time forecasting with minimal additional memory and computation.

In general, we have the following contributions.
(1) We formulate the scalable customer-level load forecasting problem for distribution feeders.
(2) We propose M$^2$OE$^2$-GL, a global-to-local extension of our M$^2$OE$^2$, which brings a highly efficient and adaptable forecaster family.
(3) We identify, justify, and validate which parameter blocks to adapt, and through extensive experiments on real feeder-level data from our industry partners, show that proper fine-tuning achieves substantial error reductions.

\section{Problem Formulation}

We consider the problem of scalable probabilistic load forecasting for massive numbers of heterogeneous customer groups under realistic storage and computation constraints.
\begin{itemize}
    \item \textbf{Given:} A collection of load and external-covariate datasets $\{\mathcal{D}_g\}_{g \in \mathcal{G}}$, where each group index $g \in \mathcal{G}$ corresponds to a customer group (e.g., residential, commercial, industrial, or different regions/feeders).
    \item \textbf{Find:} A shared global base model $f_{\boldsymbol{\theta}_0}$ with parameters $\boldsymbol{\theta}_0$, and a set of group-specific adaptation parameters $\{\boldsymbol{\phi}_g\}_{g \in \mathcal{G}}$ such that each adapted model $f_{\boldsymbol{\theta}_0,\boldsymbol{\phi}_g}$ provides accurate probabilistic forecasts for group $g$, while keeping each $\boldsymbol{\phi}_g$ compact enough to satisfy the overall storage and computation budgets for a large $\mathcal{G}$.
\end{itemize}

\section{Preliminaries}

\subsection{Meta–Mixture-of-Experts for External Data (M$^2$OE$^2$)}

In this subsection, we briefly review the cutting-edge probabilistic load forecaster M$^2$OE$^2$, and refer readers to \cite{li2025external} for more details. In essence, M$^2$OE$^2$ employs an advanced mixture-of-expert (MoE) architecture with a gating mechanism that selectively identifies the most relevant external data sources across different times and operating conditions. The resulting meta-representation is then used to modulate a carefully chosen subset of the base neural network’s parameters, so that the model focuses adaptation on the most impactful weights while keeping the overall number of modulated parameters moderate. 

Specifically,  M$^2$OE$^2$ proposes the following meta-representation:

\begin{equation}
\label{eqn:moe2}
\theta_i = \sum_{j=1}^M l_{ji} \cdot g_j(w_{ji}) + \theta_0,
\end{equation}
where $w_{ji} \in \mathcal{D}_g$ ($\forall g\in \mathcal{G}$) is the $j^{th}$ ($1\leq j\leq M$) external data at time $t_i$. $g_j(\cdot)$ is the $j^{th}$ hypernetwork, a two-layer Multilayer Perceptron (MLP) with tanh as the activation function. $l_{ji}$ is the $j^{th}$ weight of a gating neural network, and we have $\sum_{j=1}^M l_{ji}=1$. \( \theta_0 \) is a static, trainable weight matrix. Consequently, Eq. \eqref{eqn:moe2} shows that our model can either select some external data sources using the gating weights $l_{ji}$ in the MoE architecture or even give up all external data and pick $\theta_0$ in the Resnet style skip-connection \cite{he2016deep}.  

$\theta_i$ is used as an input lifting matrix for the load data $\boldsymbol{x}_i\in \mathcal{D}_g$ at time $t_i$. Essentially, we utilize the external data to determine what historical load information is used to forecast the future. Strict justifications of this design is in \cite{li2025external}.
\begin{equation}
\begin{aligned}
\label{eqn:xprime}
\boldsymbol{x}'_i &= \theta_i \boldsymbol{x}_i.
\end{aligned}
\end{equation}

The transformed feature $\boldsymbol{x}'_i$ will be treated as the input to a base deep learning model. In \cite{li2025external}, we propose to employ a sequence-to-sequence variational autoencoder (VAE), where the previous week's data is encoded to a contextual hidden feature, which is further converted to a latent variable $\boldsymbol{z}_{i+1}$ at time $t_{i+1}$, using the reparameterization trick. Subsequently, a decoder models the conditional distribution of the future load given the latent variable:
\begin{equation}
\begin{aligned}
\label{eqn:rnn_decode}
p(\boldsymbol{x}_{i+1} \mid \boldsymbol{z}_{i+1}) &= \mathcal{N}\bigg(\boldsymbol{\mu}_x(\boldsymbol{z}_{i+1}), \operatorname{diag}\big(\boldsymbol{\sigma}_x^2(\boldsymbol{z}_{i+1})\big)\bigg),\\
\boldsymbol{\mu}_x(\boldsymbol{z}_{i+1}) &= \rho(W_{\mu_x z} \boldsymbol{z}_{i+1} + \boldsymbol{b}_{\mu x}),\\
\log \boldsymbol{\sigma}_x^2(\boldsymbol{z}_{i+1}) &= \rho(W_{\sigma_x z} \boldsymbol{z}_{i+1} + \boldsymbol{b}_{\sigma x}),
\end{aligned}
\end{equation} 
where $\boldsymbol{\mu}_x(\cdot)$ and $\boldsymbol{\sigma}_x(\cdot)$ are the output head layers to predict the mean and the variance of future loads. $W_{\mu_x z}$, $W_{\sigma_x z}$, $\boldsymbol{b}_{\mu x}$, and $\boldsymbol{b}_{\sigma x}$ are the weight and bias terms of these layers. $\rho(\cdot)$ is the activation function. In summary, the decoder, conditioned on all observed load and external data up to the current week (including the previous week), produces the mean and variance estimates for the future loads.

The training objective is to minimize the negative Evidence Lower Bound (ELBO):

\begin{equation}
\begin{aligned}
\label{eqn:rnn_elbo}
\mathcal{L}_{\text{ELBO}} = &-\sum_{i=1} \mathbb{E}_{q(\boldsymbol{z}_{i+1} \mid \boldsymbol{x}_i,\boldsymbol{h}_{i-1})} \big[ \log p(\boldsymbol{x}_{i+1} \mid \boldsymbol{z}_{i+1}) \big] \\
&+ \lambda\cdot\text{KL}\big(q(\boldsymbol{z}_{i+1} \mid \boldsymbol{x}_i,\boldsymbol{h}_{i-1}) \| p(\boldsymbol{z}_{i+1})\big),
\end{aligned}
\end{equation}
where the first term is the negative log-likelihood, and the second term is the Kullback–Leibler (KL) divergence between the posterior of the latent distribution for $\boldsymbol{z}_{i+1}$ and the prior standard Gaussian distribution $p(\boldsymbol{z}_{i+1}) = \mathcal{N}(\boldsymbol{0}, \boldsymbol{I})$, which regularizes the latent space and avoids overfitting. $\lambda$ is a positive weight.

\subsection{Low-rank Adaptation for Fine-tuning}

Low-rank adaptation (LoRA) is a parameter-efficient fine-tuning technique that avoids updating all entries of large weight matrices. Instead of directly optimizing a full matrix $W \in \mathbb{R}^{d_{\text{out}} \times d_{\text{in}}}$, LoRA keeps $W$ frozen and parameterizes its update as a low-rank decomposition. Concretely, the adapted weight is written as
\begin{equation}
\label{eqn:lora_update}
    W' = W + \Delta W,
\end{equation}
where the update $\Delta W$ is expressed as
\begin{equation}
\label{eqn:lora_decompose}
    \Delta W = \frac{\alpha}{r}\,AB, \quad 
    A \in \mathbb{R}^{d_{\text{out}} \times r},~ 
    B \in \mathbb{R}^{r \times d_{\text{in}}},
\end{equation}
and $\alpha$ is a scaling factor. During fine-tuning, only the low-rank factors $A$ and $B$ are updated while $W$ remains fixed, which greatly reduces the number of trainable parameters and allows many task- or group-specific adapters to be stored efficiently on top of a shared backbone model.

\section{Proposed Framework}

In this section, we describe our three-phase framework to extend M$^2$OE$^2$ to M$^2$OE$^2$-GL from a global model to local models, producing a scalable family of forecasters from massive load/external datasets $\{\mathcal{D}_g\}_{g \in \mathcal{G}}$. Phase~1 performs global pre-training of a shared base model using all groups; Phase~2 introduces low-rank group-wise adaptation; and Phase~3 deploys the resulting family of forecasters for online prediction. Below, we detail Phase~1.

\subsection{Phase 1: Global Pre-training with M$^2$OE$^2$}

In the first phase, we employ the Meta--Mixture-of-Experts for External Data (M$^2$OE$^2$) as a high-capacity probabilistic base model and train it on the aggregated data $\{\mathcal{D}_g\}_{g \in \mathcal{G}}$ to obtain a shared backbone $f_{\boldsymbol{\theta}_0}$. According to the objective in Eq. \eqref{eqn:rnn_elbo} for M$^2$OE$^2$, we need to solve
\[
    \boldsymbol{\theta}_0 
    ~=~ \arg\min_{\boldsymbol{\theta}} \, \mathcal{L}_{\mathrm{ELBO}}\big(\boldsymbol{\theta}; \{\mathcal{D}_g\}_{g \in \mathcal{G}}\big),
\]
using standard stochastic gradient-based optimization. Concretely, at each training step $k$, we sample mini-batches from the union of all groups and update the parameters via
\[
    \boldsymbol{\theta}^{(k+1)} 
    ~=~ \boldsymbol{\theta}^{(k)} 
    - \eta \,\nabla_{\boldsymbol{\theta}} 
    \mathcal{L}_{\mathrm{ELBO}}\big(\boldsymbol{\theta}^{(k)}; \mathcal{B}^{(k)}\big),
\]
where $\eta$ is the learning rate, and $\mathcal{B}^{(k)} \subset \bigcup_{g \in \mathcal{G}} \mathcal{D}_g$ is the mini-batch at step $k$. After convergence, we obtain the global pre-trained backbone $f_{\boldsymbol{\theta}_0}$, which captures common relationships between loads and external covariates across all customer groups and serves as the starting point for subsequent group-wise adaptation.

\subsection{Phase 2: Low-rank Adaptation of Output Heads}

In the second phase, we introduce group-wise adaptation on top of the global backbone \(f_{\theta_0}\). Our design choice to \textbf{adapt only the output heads in Eq. \eqref{eqn:rnn_decode}} is driven by both modeling and scalability considerations. First, M$^2$OE$^2$ is highly capable of integrating useful information from past loads and external covariates, even when they exhibit different distributions across groups: (i) its MoE with gating and meta-representation jointly route inputs to context-appropriate experts and modulate a rich subspace of backbone parameters, yielding shared latent features that are robust to heterogeneous operating conditions; and (ii) its sequence-to-sequence variational autoencoder (VAE) structure maps diverse historical trajectories into a shared latent space, where temporal patterns and uncertainty are modeled explicitly; by learning a probabilistic latent prior, the model can absorb different distribution shifts (e.g., changes in level, variance, or shape) as variations in the latent variables rather than requiring separate feature extractors for each group. As a result, we do not impose any group-specific fine-tuning on the input or hidden feature transitions of the backbone.

In contrast, the output head in Eq. \eqref{eqn:rnn_decode}, which maps the latent representation to the predictive mean and variance of the load distribution, is directly sensitive to group-specific output statistics (e.g., level, volatility, uncertainty range). We therefore apply LoRA-based adaptation only at these output layers, allowing each group to adjust its predictive distribution without disturbing the globally useful feature extractor. This choice not only aligns with the role of the output head but also substantially reduces computation and storage: group-specific adapters remain small, so utilities can maintain many group-wise variants of the model without violating practical deployment constraints.

Empirically, we evaluate several adaptation choices (deeper layers, gating layers, and latent-projection layers) and find that updating only the output heads provides the best trade-off between accuracy and parameter efficiency. The numerical comparisons are reported in Section~\ref{sec:exp}.

Recall that M$^2$OE$^2$ uses two output projection matrices in Eq. \eqref{eqn:rnn_decode}, \(W_{\mu_x z}\) and \(W_{\sigma_x z}\). Let \(W_{\mu_x z}^{(0)}\) and \(W_{\sigma_x z}^{(0)}\) denote the corresponding weights in the pre-trained backbone \(\theta_0\). For each group \(g \in \mathcal{G}\), we define adapted output heads via LoRA-style low-rank updates:
\begin{equation}
\label{eqn:lora_moe}
\begin{aligned}
    W_{\mu_x z}^{(g)} 
    &= W_{\mu_x z}^{(0)} + \Delta W_{\mu_x z}^{(g)}, 
    \quad 
    \Delta W_{\mu_x z}^{(g)} = A_{\mu,g} B_{\mu,g}, \\
    W_{\sigma_x z}^{(g)} 
    &= W_{\sigma_x z}^{(0)} + \Delta W_{\sigma_x z}^{(g)}, 
    \quad 
    \Delta W_{\sigma_x z}^{(g)} = A_{\sigma,g} B_{\sigma,g},
\end{aligned}
\end{equation}
where \(A_{\mu,g}, B_{\mu,g}, A_{\sigma,g}, B_{\sigma,g}\) are group-specific low-rank factors. During Phase~2, we \emph{freeze} all backbone parameters \(\theta_0\) and only update the group-wise adaptation parameters
\[
    \phi_g := \{A_{\mu,g}, B_{\mu,g}, A_{\sigma,g}, B_{\sigma,g}\}, \quad g \in \mathcal{G},
\]
using mini-batches drawn from \(\mathcal{D}_g\). Concretely, at each fine-tuning step \(k\) for group \(g\), we perform the gradient update
\[
    \phi_g^{(k+1)} 
    = \phi_g^{(k)} 
    - \eta_{\text{ft}} \,\nabla_{\phi_g} \mathcal{L}_{\mathrm{ELBO}}\big(\theta_0, \phi_g^{(k)}; \mathcal{B}_g^{(k)}\big),
\]
where \(\eta_{\text{ft}}\) is the fine-tuning learning rate and \(\mathcal{B}_g^{(k)} \subset \mathcal{D}_g\) is the mini-batch for group \(g\) at step \(k\). This yields a family of adapted models \(f_{\theta_0,\phi_g}\) that specialize the predictive mean and variance for each group while keeping the number of additional parameters per group small enough to scale to massive numbers of customers.

\subsection{Phase 3: Online Prediction with A Family of Forecasters}

In Phase~3, we deploy the pre-trained backbone $\boldsymbol{\theta}_0$ together with the group-wise adapters $\{\boldsymbol{\phi}_g\}_{g \in \mathcal{G}}$ to form a family of specialized models. Using the LoRA-based updates in \eqref{eqn:lora_moe}, each group $g$ is served by an adapted forecaster $f_{\boldsymbol{\theta}_0,\boldsymbol{\phi}_g}$, where the shared backbone parameters $\boldsymbol{\theta}_0$ are combined with the group-specific low-rank updates $\boldsymbol{\phi}_g$ at the output heads.

At inference time, for any incoming load/external data belonging to group $g$, we route it to the corresponding model $f_{\boldsymbol{\theta}_0,\boldsymbol{\phi}_g}$, which takes the group-$g$ historical load and covariates as input and outputs the probabilistic forecasts (e.g., mean and variance, or full predictive distribution) for the future loads of group $g$. Since only the small adapters $\boldsymbol{\phi}_g$ differ across groups and the backbone $\boldsymbol{\theta}_0$ is shared, online prediction remains computationally efficient and easily scalable to massive numbers of customer groups.

\section{Experiment}
\label{sec:exp}
\subsection{Settings}

\begin{figure}[htbp]
\centering
\includegraphics[width=1\columnwidth]{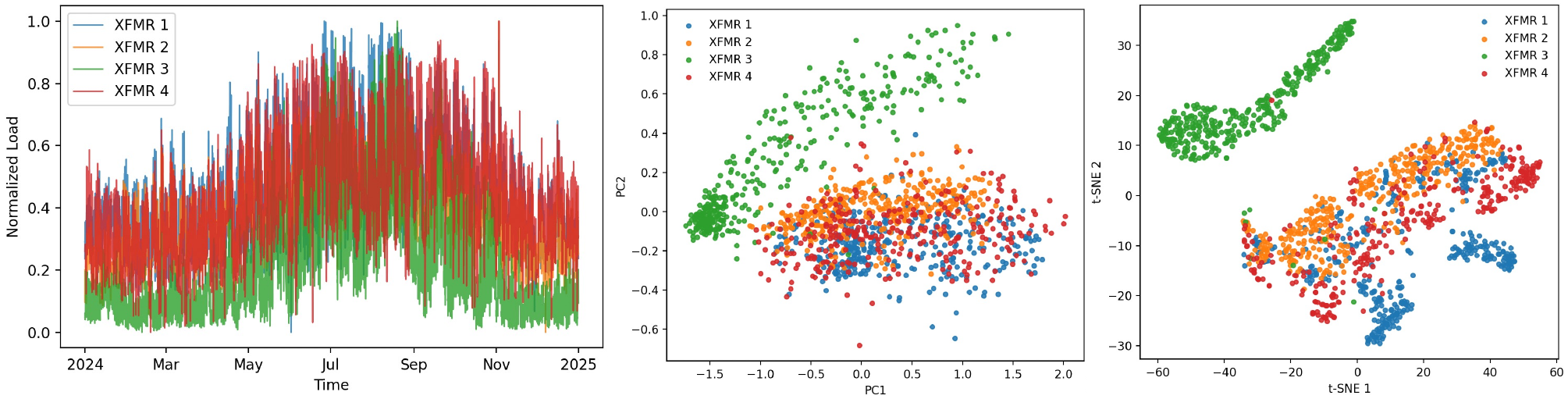}
\caption{
Transformer loads from our industrial partner: Residential (green), small commercial (orange and red), and large commercial (blue). The left figure is the raw time series in a year, and the middle and right figures are the 2-D compressed data of a day's time series using PCA and t-SNE, respectively.
}
\label{fig:Oncor_Transformer_vis}
\end{figure}

Fig. \ref{fig:Oncor_Transformer_vis} shows three load data groups from our industrial partners: residential (green), small commercial (orange and red), and large commercial (blue). The left figure shows the raw data visualization, while the middle and right figures show 2-D visualizations in the feature space, where each point represents a daily time series compressed by Principal Component Analysis (PCA) and t-distributed Stochastic Neighbor Embedding (t-SNE) \cite{maaten2008visualizing}, respectively. Obviously, the residential loads have high distribution shifts to the commercial loads. In the following test, we will use all the data to pretrain the model, and use the residential data to conduct fine-tuning and evaluations.  We introduce the following benchmark methods, including CNN-GRU, RNN, and LSTM. They also contain the pre-training (Base models) and fine-tuning (GL models). We utilize the mean square error (MSE), the continuous ranked probability score (CRPS), the quantile loss, and the Winkler Score to evaluate these methods.

\subsection{General Forecasting Results}
\begin{figure}[htbp]
\centering
\includegraphics[width=\columnwidth]{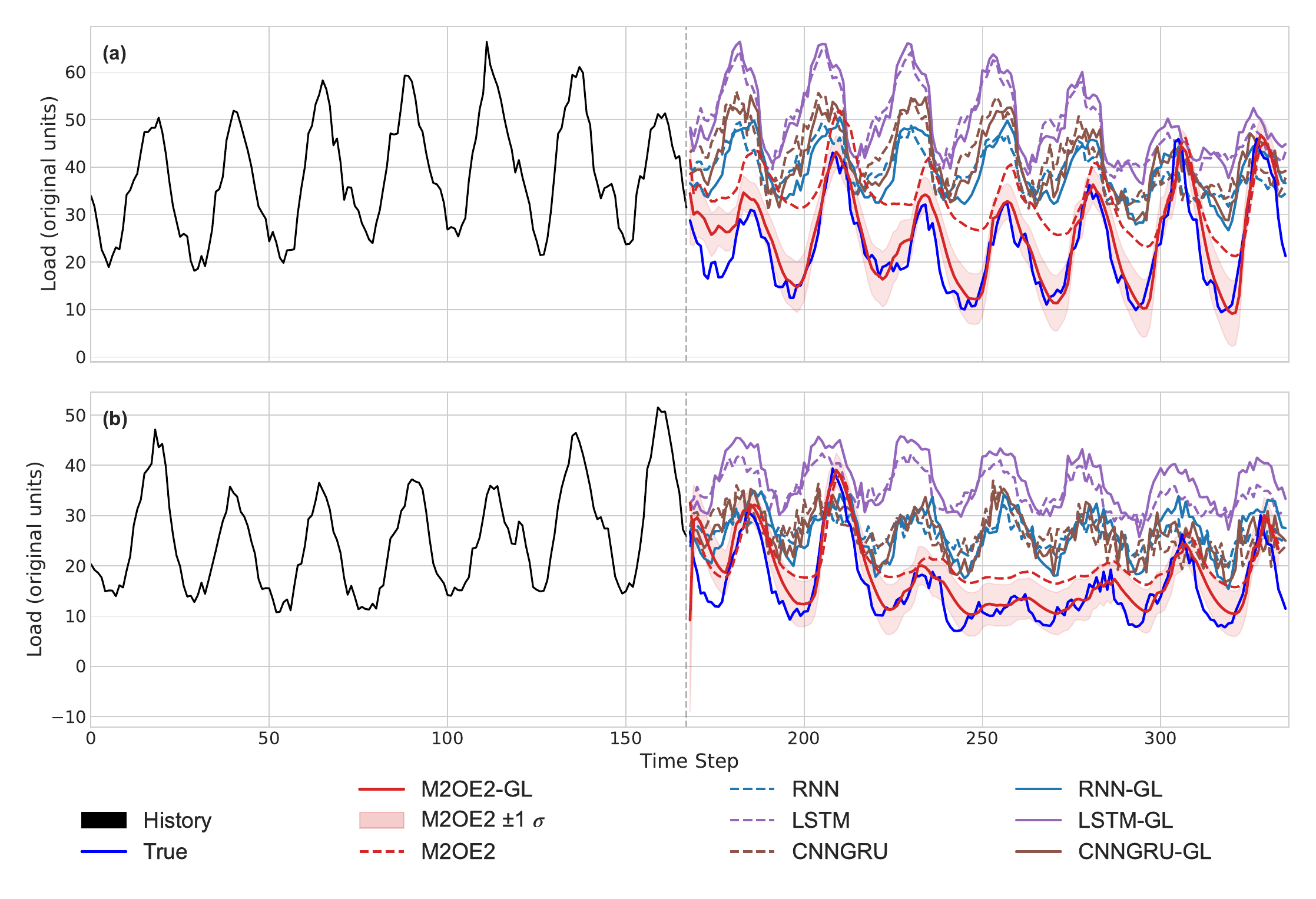}
\caption{
Load forecasting results for M$^2$OE$^2$ and baseline models on two test trajectories. The plots compare the ground truth load (blue line) against the mean prediction of our fine-tuned M$^2$OE$^2$-GL (Ours, red line), our pre-trained M$^2$OE$^2$-base (red dotted line), and other baseline models (RNN, LSTM, CNNGRU). The shaded red area represents the $\pm 1\sigma$ uncertainty bounds predicted only by the M$^2$OE$^2$-GL.
}
\label{fig:allmodel}
\end{figure}

We report deterministic and probabilistic metrics in Table~\ref{tab:model_comparison_final}, with visual comparisons in Fig.~\ref{fig:allmodel}. The base M$^2$OE$^2$ attains errors that are \emph{10--20\% of} those of strong baselines, reflecting its capacity and efficiency. Building on this, our global-to-local (GL) fine-tuned variant delivers an additional \emph{30--50\%} error reduction relative to the base model, highlighting the effectiveness of the LoRA-based updates to the output matrices in Eq.~\eqref{eqn:lora_moe}. Fig.~\ref{fig:allmodel} visually illustrates these gains.

\begin{table}[H]
\centering
\begin{minipage}{\columnwidth}

\caption{Performance metrics ($\times 10^{-2}$) for different methods.}
\label{tab:model_comparison_final}

\resizebox{\linewidth}{!}{%
\begin{tabular}{lrrrrrrrr}
\toprule
Model & \multicolumn{2}{c}{M$^2$OE$^2$} & \multicolumn{2}{c}{LSTM} & \multicolumn{2}{c}{RNN} & \multicolumn{2}{c}{CNN-GRU} \\
\cmidrule(r){2-3} \cmidrule(r){4-5} \cmidrule(r){6-7} \cmidrule(r){8-9}
Method &    GL &    Base &     GL &     Base &    GL &     Base &    GL &     Base \\
\midrule
MSE          &  \textbf{0.34} &  0.71 &   2.62 &   2.83 &  0.81 &   1.09 &  1.51 &   1.83 \\
CRPS         &  \textbf{3.01} &  4.96 &   9.35 &  10.52 &  5.67 &   6.88 &  7.04 &   8.09 \\
QuantileLoss &  \textbf{1.32} &  2.89 &   4.51 &   5.09 &  2.71 &   3.32 &  3.35 &   3.88 \\
WinklerScore & \textbf{21.02} & 54.36 & 138.79 & 163.27 & 79.47 & 104.09 & 95.25 & 117.24 \\
\bottomrule
\end{tabular}%
} 

\end{minipage} 
\end{table}

\subsection{Ablation Study: Which Layer Should Be Adapted?}
To identify which layers benefit most from LoRA, we apply adapters separately to the input, hidden-transition, and output matrices. Table~\ref{tab:lora_ablation_final} summarizes the results. We find that fine-tuning earlier layers (input or hidden) yields limited gains and can even underperform the base model. This reveals that early layers must encode globally shared representations, which should not be restricted to specific sub-datasets. 

\begin{table}[H] 
\centering

\caption{Results of fine-tuning on different weight matrices.}
\label{tab:lora_ablation_final}
\small
\setlength{\tabcolsep}{5pt}

\begin{tabular}{lcccc} 
\toprule
Target Layer & MSE & CRPS & QLoss & WScore \\
\midrule

Input matrix & 0.79 & 4.97 & 3.12 & 54.55 \\
Hidden transition matrix & 0.59 & 3.65 & 2.53 & 50.21 \\
Output matrix & \textbf{0.34} & \textbf{3.01} & \textbf{1.32} & \textbf{21.02} \\
\bottomrule
\end{tabular}
\end{table}

\subsection{Sensitivity Analysis: What is the Adaptation Size?}

We further perform a rank ablation and find that with $r=8$ (see Eq.~\ref{eqn:lora_decompose}), M$^2$OE$^2$-GL attains the best performance, indicating an effective accuracy–efficiency trade-off with minimal storage overhead.

\begin{figure}[htbp]
\centering
\includegraphics[width=0.9\columnwidth]{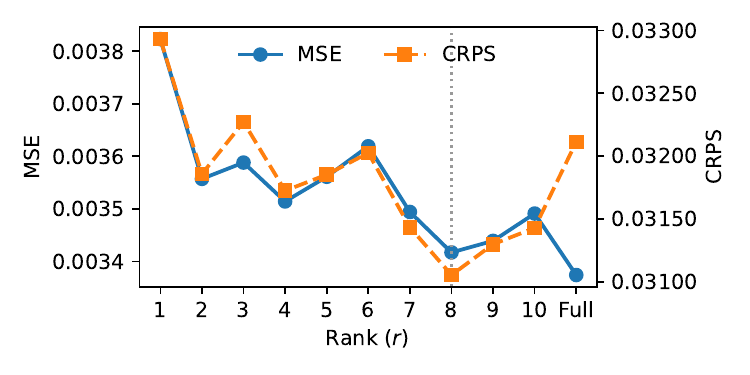}
\caption{
The figure conducts a sensitivity test on the LoRA rank(r) on fine-tuning performance. It compares LoRA models with r varying from 1 to 10 against a full-rank ('Full') fine-tuning model based on our two metrics. 
}
\label{fig:Sensitivity Analysis}
\end{figure}

\section{Conclusions}
We present M$^2$OE$^2$-GL, a fine-tuned deep learning family for adapted load forecasting across massive numbers of customers. By applying LoRA-based adaptation to the most impactful layers in M$^2$OE$^2$, the framework achieves an effective tradeoff between accuracy, efficiency, and model complexity, making it highly suitable for deployment in large-scale distribution feeders. 


\bibliographystyle{IEEEtran}
\bibliography{IEEEabrv,reference}

\end{document}